\documentclass[10pt,twocolumn,a4paper]{article}
\usepackage{algorithm2e}
\usepackage{amssymb}
\usepackage{bbm}
\usepackage{dsfont}
\usepackage{amsmath}
\usepackage{fullpage}
\usepackage{subcaption}
\usepackage[font=sf,labelfont={sf,bf}, margin=1.5cm]{caption}
\usepackage[pdftex]{graphicx}
\usepackage[font=small,labelfont=bf]{caption}
\usepackage[hmargin=3cm,vmargin=3.5cm]{geometry}
\usepackage{abstract}
\usepackage{wrapfig}
\usepackage{float}
\usepackage{graphicx}
\usepackage{hyperref}
\usepackage{tikz}

\newcommand{\HRule}{\rule{\linewidth}{0.5mm}}

\title{Gaussian Process Optimization}

\begin{document}

\begin{minipage}{\textwidth}
\begin{center}       
\HRule \\[0.4cm]
{ \large \bfseries A Hybrid Monte Carlo Architecture for Parameter Optimization}\\[0.2cm]
\HRule \\[1.5cm]

\hspace{-.3cm}
\begin{minipage}{0.585\textwidth}
\begin{flushleft} \small
\textbf{Author: James Brofos\\}
\emph{\href{mailto:james.a.brofos.15@dartmouth.edu}{james.a.brofos.15@dartmouth.edu}\\Dartmouth College}
\end{flushleft}
\end{minipage}
\begin{minipage}{0.4\textwidth}
\begin{flushleft} \small
\textbf{Advisor: Meifang Chu\\}
\emph{Lecturer in Mathematical Finance\\Dartmouth College}
\end{flushleft}
\end{minipage}
\end{center}

\vspace{-2.3cm}
\hspace{11.1cm}
\textbf{\small \today}

\vspace{2.3cm}

\begin{abstract}
\noindent
Much recent research has been conducted in the area of Bayesian learning, particularly with regard to the optimization of hyper-parameters via Gaussian process regression \cite{bayesian,expected}. The methodologies rely chiefly on the method of maximizing the expected improvement of a score function with respect to adjustments in the hyper-parameters. In this work, we present a novel algorithm that exploits notions of confidence intervals and uncertainties to enable the discovery of the best optimal within a targeted region of the parameter space. We demonstrate the efficacy of our algorithm with respect to machine learning problems and show cases where our algorithm is competitive with the method of maximizing expected improvement.
\end{abstract}

\end{minipage}

\section{Introduction}
\small
We begin by formally defining the notion of the Gaussian process as prior distribution on functions $f$ where $f: \Theta \to \mathbb{R}$ and we consider $\Theta$ to be a ``state-space'' of parameters. In particular, given tuples $\left(\theta,y\right) \in \Theta \times \mathbb{R}$, we assume that $y \sim \mathcal{N}\left(f\left(\theta\right),\sigma^2\right)$. We say that a series of such tuples, of cardinality $n$, induces a multivariate Gaussian distribution in $\mathbb{R}^n$. This Gaussian architecture is appealing for several reasons:

\begin{enumerate}
\item It elegantly fits a basis function to the data, allowing for trivial inference of the behavior of all points in $\Theta$.
\item The underlying Gaussian assumptions permit statistical notions of \emph{expected improvement} and \emph{uncertainty} to arise in closed form from the fitted model.
\item The prior two points lead naturally to a framework that enables Gaussian processes to optimize parameters in machine learning models via a principled search of $\Theta$.
\end{enumerate}

Optimization frameworks of this form offer an immediate advantage over discrete parameter optimization methodologies such as $k$-fold cross-validation, which requires $k^m$ performance evaluations of the learned model if $m$ is the cardinality of the discrete set. This is computationally expensive and fails to generate knowledge of the model's performance for $\theta \in \Theta$ when $\theta$ is not a member of the discrete parameter set used by cross-validation. \textcolor{white}{.... .. .... .. .... .. ..... ..  ..... .. .... ..... ..... .. ...... ....... $~$}

\vspace{9.74cm}

Current generation Gaussian process optimization methods exploit the the expected improvement of the model performance above the current best at all points in $\Theta$. The improvement at $\theta^*$ is $\mathcal{I}\left(\theta^*\right) = f\left(\theta^*\right) - y_{\text{best}}$, where $y_{\text{best}}$ is the current best score of the objective function. It can be shown \cite{expderivation} that the expected improvement is:
\begin{align}
\mathbb{E}\left[\mathcal{I}\left(\theta^*\right)\right] = \max \left\{0,\sigma\left(\theta^*\right)\left[u\Phi\left(u\right) + \phi\left(u\right)\right]\right\}
\end{align}
Here we represent the standard deviation of the prediction at $\theta^*$ as $\sigma\left(\theta^*\right)$ and let $u = \frac{f\left(\theta^*\right) - y_{\text{best}}}{\sigma\left(\theta^*\right)}$. We also denote the CDF of the standard normal distribution as $\Phi\left(\cdot\right)$ and similarly for the standard normal PDF, $\phi\left(\cdot\right)$. The essential idea of these optimization algorithms is to pursue function evaluations at those points yielding highest expected improvement in the objective function, thereby extracting more information about the nature of the true, underlying objective function itself. This process is continued until no further function evaluations are expected to yield improvements.

In this work we consider additionally the applications of the \emph{probability of improvement} to enhancing the Monte Carlo nature of our algorithm. The probability of improvement can be derived as:
\begin{align}
\mathbb{P}\left[y_{\theta^*} > y_{\text{best}}\right] = \mathbb{P}\left[X < \frac{f\left(\theta^*\right) - y_{\text{best}}}{\sigma\left(\theta^*\right)}\right] \\ =
\mathbb{P}\left[X < u\right]= \Phi\left(u\right)
\end{align}
Here we say that $X \sim\mathcal{N}\left(0,1\right)$, which is trivially shown as true. For the purposes of this work, we will refer to the expected improvement, probability of improvement, and mean-value criteria for point selection as \emph{anticipation equations}.

\subsection{Squared Exponential Covariance Function}
\small

Equally necessary to the definition of the Gaussian process is the covariance kernel, which permits the Gaussian process to express a versatile set of basis functions to fit the underlying objective function. A common choice of kernel is \emph{squared exponential}, which defines a matrix $\mathbf{C}$:
\begin{align}
\mathbf{C}\left(\theta_i,\theta_j\right) = \alpha \exp \left[-\frac{1}{2}\sum_{d=1}^D \frac{\left(\theta_i^{(d)} - \theta_j^{(d)}\right)^2}{2\gamma_d^2}\right],
\end{align}
and a vector $\mathbf{k} = \mathbf{C}\left(\theta,\theta_i\right)$. The hyper-parameters $\{\alpha,\gamma_1,\ldots,\gamma_D\}$ of the Gaussian process may be learned via maximum likelihood estimation by maximizing the \emph{evidence} of the fitted model. These are functionally related to the prediction and variance of the prediction as follows:
\begin{align}
f\left(\theta\right) = \mathbf{k}^T \mathbf{C}^{-1} \mathbf{y}\\\sigma^2\left(\theta\right) = \mathbf{C}\left(\theta,\theta\right) - \mathbf{k}^T \mathbf{C}^{-1} \mathbf{k}
\end{align}

The squared exponential kernel is a frequent choice within the Gaussian process literature, so we select it here as a practical (and interpretable) baseline for our proposed methodology. Some authors have criticized this choice of kernel as providing an unreasonably smooth interpolation of the basis \cite{bayesian}. The alternative option is that of Snook et al. though we do not implement the Mat\'{e}rn$_{\frac{5}{2}}$ kernel:
\begin{align}
\mathbf{C}\left(\theta_i,\theta_j\right) = \alpha\left(1 + \sqrt{5\Gamma} +\frac{5}{3}\Gamma\right)\exp \left(-\sqrt{5\Gamma}\right)\\\Gamma=\Gamma\left(\theta_i,\theta_j\right) = \sum_{d=1}^D \frac{\left(\theta_i^{(d)} - \theta_j^{(d)}\right)^2}{2\gamma_d^2}
\end{align}

\section{Machine Learning Problem Formalism}
\small

In the context of machine learning with are typically presented with a model $M$, which is a function of the observations $\vec{x}_i$, the targets $y_i$, and the model parameters $\theta$. The efficacy of this model can then be evaluated by a score function $\Psi\left(M\right)$, which is most commonly either the accuracy (to be maximized) or the error (to be minimized). Because $\vec{x}_i$ and $y_i$ are fixed, the ability of the model to generate predictions depends necessarily on $\theta$ (and perhaps also on random starting conditions in, for example, neural networks). Regardless of whether or not the model parameters are discrete\footnote{For example, consider the number of trees grown in a decision forest.} or continuous\footnote{For example, consider the $\sigma^2$ parameter in a SVM with a Gaussian kernel.}, it is possible to fit a regression function through the score function values $\Psi\left(M_{\theta^*}\right)$ at the point $\theta^*$.

Using this architecture, the fundamental optimization procedure is as follows: 
\vspace{1mm}
\vspace{1mm}
\noindent
\textbf{Algorithm 1: Original Gaussian Process Optimization}
\vspace{1mm}
\vspace{1mm}\noindent
\textbf{Input:} A labeled data set\\$\{\left(\vec{x}_1,y_1\right),\ldots,\left(\vec{x}_m,y_m\right)\}$ and parameters $\theta_0 \in \Theta$.

\noindent
\textbf{Output:} Proposed best parameters $\theta_{\text{best}}$ which maximize the score function. 

\noindent
\textbf{Algorithm:} Learn $M_{\theta_0}$ using the data and $\theta_0$. 

\noindent
Evaluate $\Psi\left(M_{\theta_0}\right)$ and initialize set of tuples $\{\left(\theta_i,\Psi\left(M_{\theta_i}\right)\right)\}$ with $\left(\theta_0,\Psi\left(M_{\theta_0}\right)\right)$.

\noindent
Initialize $\theta_{\text{best}} = \theta_0$.

\noindent
\textbf{While:} Stopping criterion \textbf{False}

\indent Fit a Gaussian process to $\{\left(\theta_i,\Psi\left(M_{\theta_i}\right)\right)\}~\forall~i$.
\indent Infer a $\theta^* \in\Theta$ that is anticipated to yield the \indent greatest difference $\Psi\left(M_{\theta^*}\right) - \Psi\left(M_{\theta_{\text{best}}}\right)$ by an \indent anticipation equation.

\indent Evaluate $\Psi\left(M_{\theta^*}\right)$ and add tuple $\left(\theta^*,\Psi\left(M_{\theta^*}\right)\right)$ \indent to $\{ \left(\theta_i,\Psi\left(M_{\theta_i}\right)\right)\}$.

\indent \textbf{If:} $\Psi\left(M_{\theta^*}\right) > \Psi\left(M_{\theta_{\text{best}}}\right)$

\indent \indent $\theta_{\text{best}} = \theta^*$

\noindent
\textbf{Return:} $\theta_{\text{best}}$
\vspace{1mm}
\vspace{1mm}

The weakness of this algorithm is that, under most circumstances, if there is \emph{no} indication that a scoring function evaluation at $\theta^*$ will lead to improvement, that point will not be evaluated. This is true even when the Gaussian process knows very little about the nature of the function at $\theta^*$. As a result, this optimization procedure can be prone to finding poor local minima due to, for instance, bad initialization of $\theta_0$. This can be combatted to an extent by pursuing multiple random starts of the algorithm, however that process begins to resemble precisely the kind of cross-validation procedure we wished to avoid.  

In the algorithm, we indicate an unspecified stopping criterion for the optimization. In our experiments, we specify that the algorithm should complete a predetermined number steps unless it converges to a maximum (either local or global) of its own accord and suspects that no further function evaluations are worthwhile, in which case termination is immediate.

\begin{figure}[ht]
  \captionsetup{font=small}
  \centering
  \begin{subfigure}[b]{.23\textwidth}
    \includegraphics[width=\textwidth]{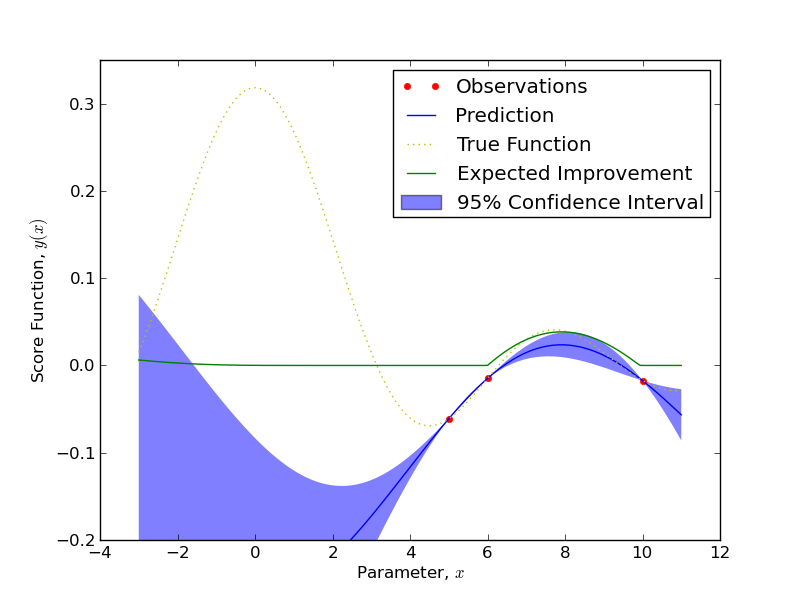}
    \caption[8pt]{Initial position for both the original and the Monte Carlo algorithms.}
  \end{subfigure}
  ~
  \begin{subfigure}[b]{.23\textwidth}
    \includegraphics[width=\textwidth]{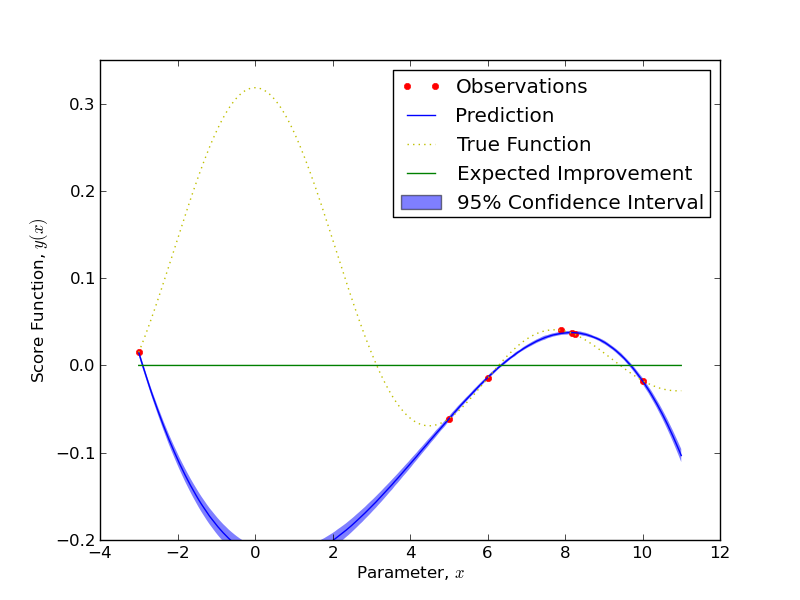}
    \caption[8pt]{Original terminates at five iterations and discovers local maximum.}
  \end{subfigure}
  \caption[8pt]{Demonstration of the enhanced Monte Carlo-based algorithm versus the original expected improvement Gaussian process optimization procedure. Notice that the enhanced algorithm finds the global optimum of the score function, whereas the alternative does not.}
  \label{fig:toy}
\end{figure}
\section{A Hybrid Optimization Algorithm}
\small

It is apparent that it would be preferable if our optimization algorithm incorporated in itself a mechanism to search for maxima in regions about which the Gaussian process can infer very little. However, it is equally apparent that an algorithm that searches only in those low-knowledge regions will be inefficient. Therefore, a superior algorithm would choose to evaluate regions of high uncertainty only a small fraction of the time, and would otherwise devote its attention to maximizing the scoring function in the typical fashion. To this end, we propose to incorporate what nearly amounts to a Metropolis-Hastings-like step such that the algorithm will use biased ``coin flips'' to determine whether or not an uncertain region is evaluated in the next iteration. 


We note that exploring the region of highest uncertainty offers an immediate advantage over other common, uncertainty-based approaches, namely the method of searching the Gaussian process' upper confidence bound. In particular, the upper confidence bound would require the additional tuning of a width parameter $\omega$. We can begin to express this idea in the following algorithm, which preserves the core of the Gaussian process optimization algorithm, yet incorporates a kind of exploratory awareness that can lead to gains.

\vspace{1mm}
\vspace{1mm}
\noindent
\textbf{Algorithm 2: Hybrid Gaussian Process Optimization}
\vspace{1mm}
\vspace{1mm} \noindent
\textbf{Input:} Labeled data set\\$\{\left(\vec{x}_1,y_1\right),\ldots,\left(\vec{x}_m,y_m\right)\}$ and parameters $\theta_0 \in \Theta$.

\begin{figure}[ht]
  \begin{subfigure}[b]{.23\textwidth}
    \includegraphics[width=\textwidth]{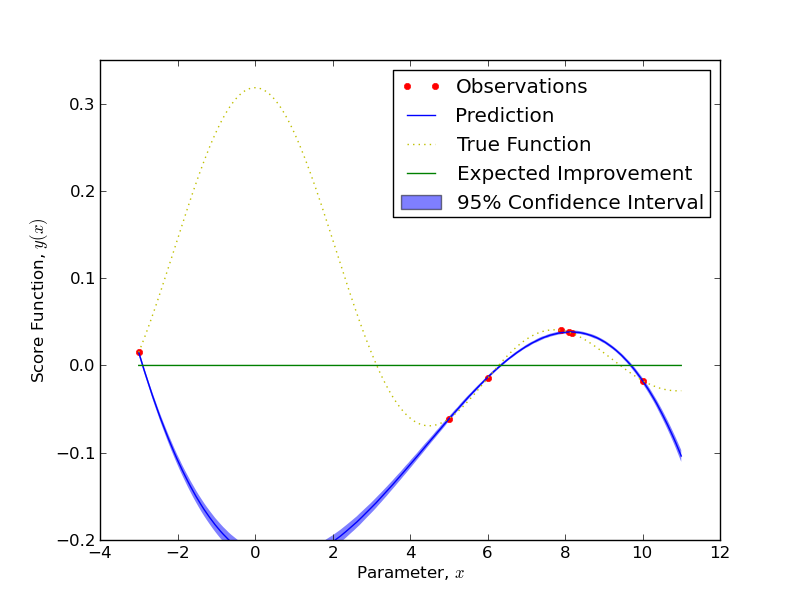}
    \caption[8pt]{Monte Carlo variant finds local maximum identically to original.}
  \end{subfigure}
  ~
  \begin{subfigure}[b]{.23\textwidth}
    \includegraphics[width=\textwidth]{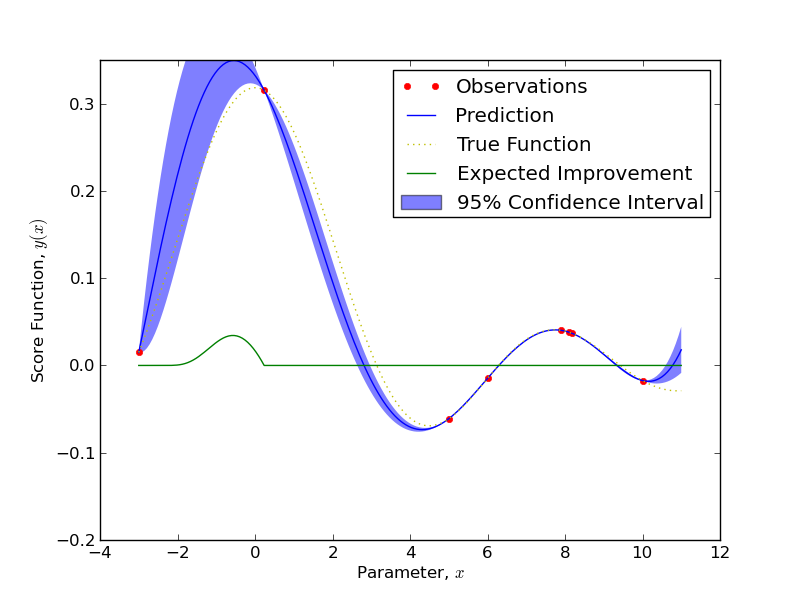}
    \caption[8pt]{Monte Carlo investigates high-uncertainty area and finds optimum.}
  \end{subfigure}
\end{figure}

\noindent
\textbf{Output:} Proposed best parameters $\theta_{\text{best}}$ which maximize the score function.

\noindent
\textbf{Algorithm:} Learn $M_{\theta_0}$ on the data and $\theta_0$. 

\noindent
Evaluate $\Psi\left(M_{\theta_0}\right)$ and initialize set of tuples $\{\left(\theta_i,\Psi\left(M_{\theta_i}\right)\right)\}$ with $\left(\theta_0,\Psi\left(M_{\theta_0}\right)\right)$.

\noindent
Initialize $\theta_{\text{best}} = \theta_0$ and set a threshold $\tau \in \left[0,1\right]$.

\noindent
\textbf{While:} Stopping criterion \textbf{False}

\indent Fit a Gaussian process to $\{\left(\theta_i,\Psi\left(M_{\theta_i}\right)\right)\}~\forall~i$.

\indent Infer a $\theta^* \in\Theta$ that is anticipated to yield the \indent greatest difference $\Psi\left(M_{\theta^*}\right)-\Psi\left(M_{\theta_{\text{best}}}\right)$ by an \indent anticipation equation.

\indent
Obtain the closed-form standard deviations of all \indent points in $\Theta$ and retrieve that point 

\indent
$\theta^{\text{u}} = \text{argmax}~\left(\sqrt{\sigma^2\left(\theta\right)}\right)~\forall~\theta\in\Theta$.

\indent Generate a random uniform value $\rho \in\left(0,1\right)$.

\indent \textbf{If:} $\rho < \tau$

\indent \indent Evaluate $\Psi\left(M_{\theta^*}\right)$ and add tuple $\left(\theta^*,\Psi\left(M_{\theta^*}\right)\right)$ \indent\indent to $\{\left(\theta_i,\Psi\left(M_{\theta_i}\right)\right)\}$

\indent\indent\textbf{If:} $\Psi\left(M_{\theta^*}\right) > \Psi\left(M_{\theta_{\text{best}}}\right)$

\indent\indent\indent $\theta_{\text{best}} = \theta^*$

\indent\textbf{Else:} 

\indent\indent Evaluate $\Psi\left(M_{\theta^{\text{u}}}\right)$ and add tuple $\left(\theta^{\text{u}},\Psi\left(M_{\theta^\text{u}}\right)\right)$ \indent\indent to $\{\left(\theta_i,\Psi\left(M_{\theta_i}\right)\right)\}$

\indent\indent\textbf{If:} $\Psi\left(M_{\theta^{\text{u}}}\right) > \Psi\left(M_{\theta_{\text{best}}}\right)$

\indent\indent\indent $\theta_{\text{best}} = \theta^{\text{u}}$

\noindent\textbf{Return:} $\theta_{\text{best}}$

\vspace{1mm}
\vspace{1mm}

In our experiments, we select the threshold $\tau = \frac{4}{5}$. In the interest of demonstrating the efficacy of our new algorithm, we construct a toy example that shows an instance where expected improvement optimization terminates before finding the global maximum, whereas our algorithm does precisely the opposite. In particular, for an input $x$, we define our score function by the equation, $y\left(x\right) = \frac{\sin \left(x\right)}{\pi x}$. We initialize both algorithms with an identical triplet of known function points, and ask the algorithms to run twenty iterations unless convergence is achieved. In the case of the original optimization algorithm, the Gaussian process quickly finds the local optimum, but chooses to discontinue searching the space after four iterations. By contrast, the hybrid architecture also finds the local optimum in four iterations, but then evaluates the point of largest uncertainty, which is near the global maximum. This phenomenon is illustrated in Figure \ref{fig:toy} and leads us to validate the hypothesis that our algorithm is capable of finding improved maxima in optimization problems. 

\begin{minipage}{\textwidth}
  \begin{table}[H]
    \centering
    \scriptsize
    \begin{tabular}{|p{3cm}|p{3cm}|p{4cm}|p{3cm}|}
      \hline
      \hline
      \multicolumn{4}{c}{Details of Experiments for the Employed Data Set}\\
      \cline{1-4}
      \emph{Domain} & \emph{Raw Features} & \emph{Response} & \emph{Data Set Cardinality}\\
      \hline
      Australian Credit Scoring & 16 & Desired credit approval of individuals based on characteristics & 690\\\hline
    \end{tabular}
    \caption{\small Data set descriptions for the experiments used to validate the efficacy of the proposed algorithm. We summarize here the domain of the application, the input features to the algorithm, the response variable we wish to predict and the number of examples provided in the data.}
  \end{table}
\end{minipage}

\subsection{Variable Threshold Algorithm}
\small

For some purposes it may be desirable not to use a fixed threshold $\tau$ for selecting a proportion of instances to search areas of high uncertainty. We therefore present an additional algorithm which incorporates a dynamic thresholding for choosing to explore low-knowledge regions. This methodology is principled in the sense that it employs the probability of improvement of the highest uncertainty point as a scaling parameter on a ``basis'' threshold $\tau'$, which may equal unity if so desired. This permits exploration of unknown spaces a portion of the time (unlike the original algorithm), yet also recognizes that it can be advantageous to focus closely on maximizing expected improvement in a fashion that is inversely proportional to the probability of improvement at the location of highest uncertainty in $\Theta$.

\vspace{1mm}
\vspace{1mm}
\noindent
\textbf{Algorithm 3: Variable Threshold Gaussian Process Optimization}
\vspace{1mm}
\vspace{1mm}\noindent
\textbf{Input:} A labeled data set\\$\{\left(\vec{x}_1,y_1\right),\ldots,\left(\vec{x}_m,y_m\right)\}$ and parameters $\theta_0 \in \Theta$.

\noindent
\textbf{Output:} Proposed best parameters $\theta_{\text{best}}$ which maximize the score function.

\noindent
\textbf{Algorithm:} Learn $M_{\theta_0}$ on the data and $\theta_0$.

\noindent 
Evaluate $\Psi\left(M_{\theta_0}\right)$ and initialize set of tuples $\{\left(\theta_i,\Psi\left(M_{\theta_i}\right)\right)\}$ with $\left(\theta_0,\Psi\left(M_{\theta_0}\right)\right)$. 

\noindent
Initialize $\theta_{\text{best}} = \theta_0$ and set a ``basis'' threshold $\tau \in \left[0,\infty\right)$.

\noindent
\textbf{While:} Stopping criterion \textbf{False}

\indent Fit a Gaussian process to $\{\left(\theta_i,\Psi\left(M_{\theta_i}\right)\right)\}~\forall~i$.

\indent Infer a $\theta^* \in\Theta$ that is anticipated to yield the \indent greatest difference $\Psi\left(M_{\theta^*}\right)-\Psi\left(M_{\theta_{\text{best}}}\right)$ by an \indent anticipation equation.

\indent
Obtain the closed-form standard deviations of all \indent points in $\Theta$ and retrieve that point

\indent
$\theta^{\text{u}} = \text{argmax}~\left(\sqrt{\sigma^2\left(\theta\right)}\right)~\forall~\theta\in\Theta$.

\indent Obtain the probability of improvement for $\theta^{\text{u}}$, $\nu$ \indent and generate a random uniform value $\rho \in\left(0,1\right)$.

\indent \textbf{If:} $\rho < \nu\tau$

\indent \indent Evaluate $\Psi\left(M_{\theta^*}\right)$ and add tuple $\left(\theta^*,\Psi\left(M_{\theta^*}\right)\right)$ \indent\indent to$\{\left(\theta_i,\Psi\left(M_{\theta_i}\right)\right)\}$

\indent\indent\textbf{If:} $\Psi\left(M_{\theta^*}\right) > \Psi\left(M_{\theta_{\text{best}}}\right)$

\indent\indent\indent $\theta_{\text{best}} = \theta^*$

$~~~~~~~~~~~~~~~~~~~~~~~~~~~~$
$~~~~~~~~~~~~~~~~~~~~~~~~~~~~$
$~~~~~~~~~~~~~~~~~~~~~~~~~~~~$
$~~~~~~~~~~~~~~~~~~~~~~~~~~~~$
$~~~~~~~~~~~~~~~~~~~~~~~~~~~~$
$~~~~~~~~~~~~~~~~~~~~~~~~~~~~$
$~~~~~~~~~~~~~~~~~~~~~~~~~~~~$
$~~~~~~~~~~~~~~~~~~~~~~~~~~~~$
$~~~~~~~~~~~~~~~~~~~~~~~~~~~~$
$~~~~~~~~~~~~~~~~~~~~~~~~~~~~$
$~~~~~~~~~~~~~~~~~~~~~~~~~~~~$
$~~~~~~~~~~~~~~~~~~~~~~~~~~~~$
$~~~~~~~~~~~~~~~~~~~~~~~~~~~~$
$~~~~~~~~~~~~~~~~~~~~~~~~~~~~$
$~~~~~~~~~~~~~~~~~~~~~~~~~~~~$
$~~~~~~~~~~~~~~~~~~~~~~~~~~~~$
$~~~~~~~~~~~~~~~~~~~~~~~~~~~~$
$~~~~~~~~~~~~~~~~~~~~~~~~~~~~$
$~~~~~~~~~~~~~~~~~~~~~~~~~~~~$

$~~~~~~~~~~~~~~~~~~~~~~~~~~~~$
$~~~~~~~~~~~~~~~~~~~~~~~~~~~~$
$~~~~~~~~~~~~~~~~~~~~~~~~~~~~$
$~~~~~~~~~~~~~~~~~~~~~~~~~~~~$
$~~~~~~~~~~~~~~~~~~~~~~~~~~~~$
$~~~~~~~~~~~~~~~~~~~~~~~~~~~~$
$~~~~~~~~~~~~~~~~~~~~~~~~~~~~$

\indent\textbf{Else:}

\indent\indent Evaluate $\Psi\left(M_{\theta^{\text{u}}}\right)$ and add tuple $\left(\theta^{\text{u}},\Psi\left(M_{\theta^\text{u}}\right)\right)$ \indent\indent to $\{\left(\theta_i,\Psi\left(M_{\theta_i}\right)\right)\}$

\indent\indent\textbf{If:} $\Psi\left(M_{\theta^{\text{u}}}\right) > \Psi\left(M_{\theta_{\text{best}}}\right)$

\indent\indent\indent $\theta_{\text{best}} = \theta^{\text{u}}$

\noindent
\textbf{Return:} $\theta_{\text{best}}$
\vspace{1mm}
\vspace{1mm}

\section{Experimental Results}

We now turn our attention to analyzing the performance of the variable thresholding algorithm in application to a common machine learning benchmark. We use the Australian credit approval data set available at the UCI Machine Learning repository \cite{uci}. We summarize important statistics of this dataset in Table 1. We employ this dataset for testing the algorithm because it offers a range of variable types: continuous and categorical variables, in addition to missing values. 

For the parameter selection stage, we train a random forest, which relies on minimizing the impurity measurement in a series of binary splits. To give an intuitive idea of the random forest's approach to machine learning, we provide the following formal definition. Given a set of candidate splitting tests at a particular node in a decision tree, $S\left(\tau\right)=\left\{s_1^{(\tau)},\ldots,s_n^{(\tau)}\right\}$, we seek to split the data that is satisfies:
\begin{align}
  s^* = \underset{s\in S(\tau)}{\text{arg max}~} -\sum_{c\in \mathcal{C}} P_c^{(\tau)}\log P_c^{(\tau)} 
\end{align}
Where $P_c^{(\tau)}$ represents the class posterior probability (of class $c$) for the binary split for a data point located in the region of variable space identified as $\tau$. In the case of this optimization experiment, we will attempt to identify the setting for the number of grown trees that simultaneously maximizes prediction accuracy and minimizes computation complexity.

For credit approval classification, our algorithm correctly identifies the optimal setting of parameters within ten iterations of the algorithm, having converged by the ninth. By contrast, the original Gaussian optimization algorithm fails to identify the best number of trees to create in the forest, opting for a value far larger than is empirically shown to be 

\begin{minipage}{\textwidth}
  \begin{table}[H]
    \centering
    \scriptsize
    \begin{tabular}{|p{3cm}|p{2cm}|p{2cm}|p{2cm}|p{2cm}|}
      \hline
      \hline
      \multicolumn{5}{c}{Details of Experiments for the Variable Threshold Algorithm}\\
      \cline{1-5}
      \emph{Statistic} & \emph{Average} & \emph{Minimum} & \emph{Maximum} & \emph{Standard Deviation}\\
      \hline
      Predictive Accuracy of Random Forest & {\vspace{0mm}$85\%$} & {\vspace{0mm}$81\%$} & {\vspace{0mm}$90\%$} & {\vspace{0mm}$3.24\%$}\\\hline
      Convergence Time of Optimization Algorithm & {\vspace{0mm}$10$} & {\vspace{0mm}$7$} & {\vspace{0mm}$12$} & {\vspace{0mm}$2.2$}\\\hline
    \end{tabular}
    \caption{\small We present here some relevant statistics related to our experiments in parameter optimization. Notice that in the predictive accuracy criterion, larger values are preferable. By contrast, we have that convergence time is better for smaller values. We define as convergence time the number of iterations of the algorithm that are required to map out completely the known behavior of the accuracy function.}
  \end{table}
\end{minipage}

$~~~~~~~~~~~~~~~~~~~~~~~~~$
$~~~~~~~~~~~~~~~~~~~~~~~~~$
$~~~~~~~~~~~~~~~~~~~~~~~~~$
$~~~~~~~~~~~~~~~~~~~~~~~~~$
$~~~~~~~~~~~~~~~~~~~~~~~~~$

\noindent
necessary. The variable threshold process of parameter selection, by virtue of its exploratory capability, identifies that approximately forty decision trees are necessary to achieve maximum accuracy on unnormalized features. By contrast, the original approach terminates with a selection of 97 decision trees, a significant increase in the computation complexity of the learning algorithm. We report in Table 2. some of the statistics related to the classification results of the random forest and of the convergence of the variable threshold algorithm.

\begin{figure}[h]
  \captionsetup{font=small}
  \centering
  \begin{subfigure}[b]{.23\textwidth}
    \includegraphics[width=\textwidth]{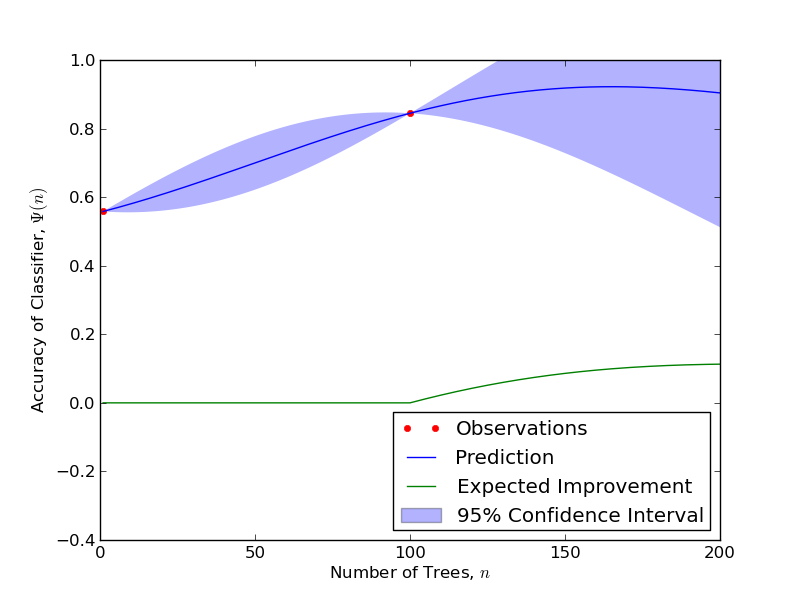}
    \caption[8pt]{Initial classification results for the original Gaussian process optimization procedure and the modified, variable thresholding approach. Notice that initially the inclusion of more trees is anticipated to improve the algorithm's predictive performance.}
  \end{subfigure}
  ~
  \begin{subfigure}[b]{.23\textwidth}
    \includegraphics[width=\textwidth]{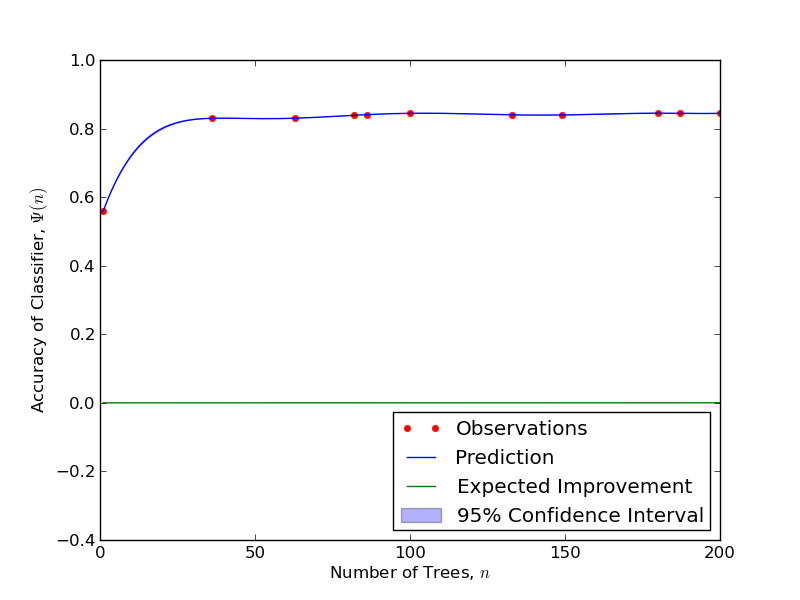}
    \caption[8pt]{The final position achieved using variable thresholding. Notice that the algorithm has identified a low-uncertainty path the correctly predicts the nature of classification for all conceivable numbers of decision trees in the random forest.}
  \end{subfigure}
\end{figure}

\section{Conclusion and Discussion}

We have presented here two novel frameworks for Gaussian process optimization. In the case of variable thresholding, we find that we are able to produce results that are superior to those yielded by the original Gaussian process approach. We believe that this particular approach to hyper-parameter value assignment has many benefits over 

$~~~~~~~~~~~~~~~~~~~~~~$
$~~~~~~~~~~~~~~~~~~~~~~$
$~~~~~~~~~~~~~~~~~~~~~~$
$~~~~~~~~~~~~~~~~~~~~~~$
$~~~~~~~~~~~~~~~~~~~~~~$
$~~~~~~~~~~~~~~~~~~~~~~$
$~~~~~~~~~~~~~~~~~~~~~~$
$~~~~~~~~~~~~~~~~~~~~~~$
$~~~~~~~~~~~~~~~~~~~~~~$
$~~~~~~~~~~~~~~~~~~~~~~$
$~~~~~~~~~~~~~~~~~~~~~~$
$~~~~~~~~~~~~~~~~~~~~~~$
$~~~~~~~~~~~~~~~~~~~~~~$
$~~~~~~~~~~~~~~~~~~~~~~$
$~~~~~~~~~~~~~~~~~~~~~~$
$~~~~~~~~~~~~~~~~~~~~~~$
$~~~~~~~~~~~~~~~~~~~~~~$
$~~~~~~~~~~~~~~~~~~~~~~$
$~~~~~~~~~~~~~~~~~~~~~~$
$~~~~~~~~~~~~~~~~~~~~~~$
$~~~~~~~~~~~~~~~~~~~~~~$
$~~~~~~~~~~~~~~~~~~~~~~$
$~~~~~~~~~~~~~~~~~~~~~~$
$~~~~~~~~~~~~~~~~~~~~~~$
$~~~~~~~~~~~~~~~~~~~~~~$
$~~~~~~~~~~~~~~~~~~~~~~$
$~~~~~~~~~~~~~~~~~~~~~~$
$~~~~~~~~~~~~~~~~~~~~~~$
$~~~~~~~~~~~~~~~~~~~~~~$
$~~~~~~~~~~~~~~~~~~~~~~$
$~~~~~~~~~~~~~~~~~~~~~~$
$~~~~~~~~~~~~~~~~~~~~~~$
$~~~~~~~~~~~~~~~~~~~~~~$
$~~~~~~~~~~~~~~~~~~~~~~$
$~~~~~~~~~~~~~~~~~~~~~~$
$~~~~~~~~~~~~~~~~~~~~~~$
$~~~~~~~~~~~~~~~~~~~~~~$
$~~~~~~~~~~~~~~~~~~~~~~$
$~~~~~~~~~~~~~~~~~~~~~~$
$~~~~~~~~~~~~~~~~~~~~~~$
$~~~~~~~~~~~~~~~~~~~~~~$
$~~~~~~~~~~~~~~~~~~~~~~$
$~~~~~~~~~~~~~~~~~~~~~~$
$~~~~~~~~~~~~~~~~~~~~~~$

\noindent
other competing techniques such as $k$-fold cross-validation. We hope that these algorithms will find application in other machine learning applications where parameter optimization is crucial. In particular, we foresee applications to neural network learning as a future application of the approach.

\end{document}